\documentclass{llncs}

\usepackage[T1]{fontenc}
\usepackage[utf8]{inputenc}
\begin{document}

\title{Extending Automatic Discourse Segmentation for Texts in Spanish to Catalan}
\titlerunning{Automatic Discourse Segmentation for Catalan}  
%
\author{Iria da Cunha\inst{1} \and Eric SanJuan\inst{2} \and Juan-Manuel Torres-Moreno\inst{2,3} \and Irene Castell\'on\inst{4} \and Marina Lloberes\inst{4}}
\authorrunning{da Cunha {\it et al.}} 
%
%
\institute{
Universidad Nacional de Educaci\'on a Distancia (UNED), Madrid, Spain\\
\email{iriad@flog.uned.es}     
           \and
           LIA, Universit\'e d'Avignon et des Pays de Vaucluse, France \\
           \email{\{juan-manuel.torres, eric.sanjuan\}@univ-avignon.fr}           
           \and
	Ecole Polytechnique de Montr\'eal - Departament de G\'enie informatique, Montr\'eal, Canada\\
	\and
	Universitat de Barcelona - Departament de Ling\" u\'istica General, Barcelona, Spain\\
	\email{icastellon@ub.edu; mllobesa8@alumnes.ub.edu}
}

\maketitle              

\begin{abstract}
At present, automatic discourse analysis is a relevant research topic in the field of NLP. However, discourse is one of the phenomena most difficult to process. Although discourse parsers have been already developed for several languages, this tool does not exist for Catalan. In order to implement this kind of parser, the first step is to develop a discourse segmenter. In this article we present  the first discourse segmenter for texts in Catalan. This segmenter is based on Rhetorical Structure Theory (RST) for Spanish, and uses lexical and syntactic information to translate rules valid for Spanish into rules for Catalan. We have evaluated the system by using a gold standard corpus including manually segmented texts and results are promising.
\end{abstract}
\section{Introduction}

Nowadays discourse parsing is a very prominent research area used in Natural Language Processing (NLP).
Recently other NLP applications and approaches that underlie discourse parsing have arose, such as Machine Translation \cite{Joty2014}, Textual Similarity \cite{daCunha2014}, and Sentiment Analysis and Opinion Mining \cite{Trnavac2014,ChenloHogenboom2013} for example. 

In order to develop these applications, discourse segmenters and parsers are needed, as well as discourse level annotated corpora. Several resources have been developed for different languages.
Regarding Iberian Peninsula's Romance languages, there are resources for Portuguese
\cite{Pardo2008} and Spanish \cite{daCunha2014}.
However, they have not been developed for Catalan yet. 
Catalan is a romance language that comes from Latin. It is spoken in several parts of Spain (mainly in Catalonia), Andorra, France (Roussillon) and Italy (Alghero), among others.
Despite this number of speakers, this is an under-resourced language and there isn't any discourse annotated corpus available.

Most of the related work mentioned in this paper relies on {\sl Rhetorical Structure Theory} (RST) by \cite{Mann1988}.
According to this theory, a text can be considered as a hierarchical tree made of elementary discourse units (EDUs) that can work as nucleus or as satellite. 
While nuclei provide relevant information about the author's point of view, satellites give additional information associated to nuclei. 
RST discourse parsing is formed by three steps: 
1) text segmentation in EDUs, 2) discourse relations analysis  and 3) discourse tree building. 

There are three strategic approaches to impulse  the development of resources for under-resourced languages: a) using the crowd and collaborative platforms; b) using technologies of interoperability with well-developed languages; and c) using Semantic Web technologies and, more specifically, Linked Data. In this work, we focus on the second approach. Therefore, the main goal of this work is to develop and evaluate the first discourse segmenter for Catalan, by adapting the existing discourse segmenter for Spanish (a technologically advanced language). 

We use the  FreeLing Shallow parser for Catalan \cite{Atserias2006} as a linguistic resource.  
The system is formed by linguistic rules based on lexical units (conjunctions and adverbs), discourse markers, syntactic structures and punctuation marks. Furthermore, this work aims to present the annotated corpus developed for evaluating the segmenter. 
This corpus has been developed as a gold standard open to scientific community.


In section \ref{sec:estadoart}, we present the state of the art on discourse segmentation.
In section \ref{sec:metodologia}, we explain the methodology used to develop our segmenter for Catalan and we go deeper into the system implementation. 
The details about the experiments and the results are described in section \ref{sec:experimentos}. 
Finally, the conclusions of the current research and the future work are presented in section \ref{sec:conclusion}.

\section{State of the Art}
\label{sec:estadoart}

As stated in \cite{Hovy2010}, discourse is one of the most difficult language levels to process automatically due to its complexity.
Actually, this difficulty makes automatic discourse analysis a challenging task because it can be applied to develop several NLP tools. In this sense, \cite{Taboada2006} and \cite{Cunha11} have made surveys upon the current research on discourse parsing and its applications. Some examples are text generation \cite{Hovy1993,Dale1992,ODonnell2001} and automatic summarization \cite{Marcu2000a,Radev2000,Pardo2002}. 
Specifically, research on discourse segmentation has been proved to be useful for different NLP tasks. For example, in \cite{molina2011,molinaarxiv2012,Molina2013} authors study the relationship between discourse segmentation and compression for sentences in Spanish. They present a method for sentence compression that uses statistical information in order to delete intra-sentence discourse segments, obtaining very good results. This kind of systems has been successfully employed for cinema or TV subtitling, and for elaboration of short messages by mobile companies, among other applications. Also, discourse segmentation has been used for machine translation. For example, \cite{Ghorbel2001} use discourse segments to align passages of texts in different languages. Usually, in this field, alignment is done between sentences, but alignment between discourse segments (that is, parts of those sentences) can offer additional information that can be useful for machine translation and other fields.

Existing discourse segmenters employ several strategies; the most productive strategy has been the use of linguistic information (mainly, lexical and syntactic information). The segmenter for English by \cite{Soricut2003} is based on a statistical model that uses lexical and syntactic features to assign a probability to the insertion of a segment boundary after every word of a sentence. The segmenter for English by \cite{Tofiloski2009} is based on linguistic rules (lexical and syntactic) and uses a constituency-based parser. The segmenter for Spanish by \cite{daCunha2010} is based on linguistic rules adapted to this language and on the grammar of the shallow parser of Freeling.

Another possible strategy to deal with discourse segmentation is machine learning. It would be the case, for example, of the segmenter for French by \cite{Afantenos2010}. However, these types of approaches need a high amount of annotated texts in order to learn and carry out an adequate segmentation. Therefore, nowadays their results are not better than the results of systems based on linguistic information.

Recently, the possibility of developing language-independent segmenters or systems using very few linguistic resources is being explored. It is the case of \cite{Saksik2013}, whose system uses general statistical techniques based on morphological tags, and general linguistic rules. The only language-dependent linguistic resource is a list of discourse markers in the language of the text. The advantages of this type of strategies are that they are easy to implement and require very few resources. Therefore, they are especially adequate for languages without NLP tools. However, currently, the results of these segmenters are not better than the results of the systems designed specifically for particular languages.

In some cases, different strategies have been applied over the same monolingual corpus and have been compared among them, with the aim of determining the most productive strategy for a specific language. It is the case of \cite{Iruskieta2013}, where both statistical and linguist strategies are applied to segment texts in Basque. Also, constituent and dependency parsing are used. In this work, the strategy based on dependency parsing is the most productive. This is due to the high complexity and particular syntactic characteristics of Basque, since it is an agglutinative language.

\section{Methodology}
\label{sec:metodologia}

We follow the linguistic strategy to develop a discourse segmenter for Catalan. There are two main reasons for applying this strategy. First, Catalan and Spanish are very similar languages, and the linguistic strategy has been applied for Spanish, obtaining good results. Second, nowadays, in general, this strategy is the most productive for all languages in the state of the art, especially if the used corpus is limited. 

\subsection{Shallow parsing}
\label{subsec:gramatica}

The used grammar is an extension of the grammar of the shallow parser for Catalan included in Freeling \cite{Atserias2006,Padro2010}.  
The aim of this extension is to detect and re-categorize those words or groups of words that can indicate a boundary between discourse segments in sentences. 
These rules indicate units that can work as discourse markers. 
For that, two lexicons of Catalan discourse markers have been processed \cite{Alonso2005,genealitat2014} and 252 markers have been obtained; they were divided in two groups: 
ambiguous markers (118) and non-ambiguous markers (134). 
Finally, the 252 rules developed were added to the grammar.

Non-ambiguous markers have been introduced in a new category `disc-mk' (discourse marker) and therefore have been re-categorized in the grammar of the shallow parser for Catalan. For example: adverbs and adverbial groups (\textit{aleshores}, `so'; \textit{aix\'i doncs}, `therefore'), prepositional groups (\textit{per causa de}, `because of') or sequences of lexical units (\textit{tot seguit}, `next'; \textit{tot i que}, `although'). On the other hand, the category `disc-mk-amb' includes composed elements that have been re-categorized from different tags of the shallow parser. For example: \textit{com a mostra} (`as it is shown by') or \textit{després} (`after').

In the case of ambiguous markers, it is necessary to take into account the context where they appear and require advance parsing. For example, the marker \textit{després} (`after') can be just an adverb (see example 1) or a discourse maker (see example 2):

\begin{enumerate}
\item \textit{Els resultats mostren que \textbf{després} del test augmentaren els valors.}  (`The results show that after the test the values increased.')
\item \textit{Els jugadors de futbol de categoria juvenil van tenir fatiga del sistema nervi\'os \textbf{després} de realitzar un test de capacitat d'esprints repetits (CER).} (`The football players of the youth category had fatigue of the nervous system after carrying out a repeated sprint test (RST).')
\end{enumerate}

In the grammar, rules related to discourse markers (both ambiguous and non-ambiguous) have priority over the other rules. 

\subsection{Implementation}
\label{sec:implementacion}

The resulting segmenter for Catalan called DiSegCAT is then generated automatically from the segmenter for Spanish.These tools are implemented in Perl, and are based on regular expressions and the Twig XML library. Firstly, DiSeg calls the FreeLing library in order to apply the discourse markers grammar. This first module transforms the syntactic tree generated by the FreeLing Shallow Parser into XML format. The second module applies the rules that detect EDUs boundaries. This module reads all the leaves of the XML syntactic tree. When it recognizes a boundary, the syntactic tree is modified by adding a new node where the boundary is located. This task is iterated twice to find sub-EDUs boundaries inside the EDUs already detected. The third DiSeg module re-reads the new XML syntactic tree to split sentences into coherent EDUs that should contain one or more verbs. 

These modules use regular expressions to recognize discourse markers, lemmas and grammatical categories. The idea was the development of a Perl script to translate the code for Spanish discourse segmentation to Catalan. This strategy presuppose that EDUs have the same grammatical structure and segmentation markers are the only elements that are altered. This assumption is  not entirely true. There are discourse markers in Spanish that correspond to one or more discourse markers in Catalan. For example, the Spanish marker \textsl{para} (`to') is \textsl{per} (`to', `by' or `through') or \textsl{en} (`in') in Catalan. Furthermore, there are grammatical categories tags, such as `vaux' for auxiliary verbs, 
that not have a correspondence in the grammar version of Catalan.
Therefore, a lexicon that transforms the Spanish DiSeg to the Catalan DiSeg cannot handle these cases but  contextual rules can contribute to solve these asymmetries between both languages.

The final implementation is the es2cat.pl perl script that translates the original DiSeg into DiSegCAT. As a consequence, any expansion of the original DiSeg is applied automatically to the Catalan version. This strategy relies on the idea that translating a NLP software is much more easier than translating texts written in natural language. Therefore, the results provided by DiSegCAT are more reliable than the original DiSeg combined with a machine translation system Catalan $\leftrightarrow$ Spanish.
 
\section{Experiments and Results}
\label{sec:experimentos}

We evaluate system performance over a corpus of manually segmented texts.
This type of evaluation  is used in previous work in this area \cite{Tofiloski2009,Afantenos2010,daCunha2010}. 

\subsection{Corpus}

The corpus includes 20 abstracts of research articles in Catalan from the medical domain, extracted from the specialized Journal of Medicine and Physical Activity and Sport \textit{Apunts: Medicina de l'esport}\footnote{http://www.raco.cat/index.php/Apunts/issue/archive}. 
This journal publishes each article in two languages, Spanish and Catalan, and provides the abstract in these two languages and also in English. This would allow us to perform experiments with parallel corpora in the future. Specifically, for this work, texts published between 2010 and 2013 were selected, in order to have recent documents. Also, texts related to different subjects were selected, such as scoliosis, attention deficit, cardiology, nutrition, etc., in order to guarantee thematic diversity. The textual genre `abstract' and the medical domain were selected to be able to compare adequately the results of our discourse segmenter for Catalan with the results obtained by the discourse segmenter for Spanish; in the evaluation of this Spanish segmenter, 20 texts with similar characteristics were used.

Once the segmentation criteria were defined for Catalan and the corpus was compiled, manual discourse segmentation was carried out. For that, two annotators were asked to segment each text of the corpus, following the mentioned criteria, individually and without questions between them, to avoid biases in the results. Both annotators are linguists and have a wide experience in corpus annotation. After the manual annotation of the 20 texts, both segmentations were compared, in order to determine the inter-annotator agreement. Annotators agreed on 264 discourse segment boundaries and they disagreed on 23 boundaries. Therefore, there was a boundary agreement of 92\%. 

Following \cite{IruskietaPHD} and \cite{IruskietaCLLT13}, we have also calculated inter-annotator agreement by using Kappa Cohen in two ways: 
taking into account words as boundaries and taking into account clauses as boundaries. For the first one, the Kappa value is 0.9556 and, for the second one (that is more conservative), the Kappa value is 0.8674. We consider that these values show that agreement between annotators is high for the task of discourse segmentation.

After this quantitative analysis, a qualitative analysis of the disagreements was done, where we observed that nearly all the disagreements were due to human mistakes.
Finally, in the line of work on this topic \cite{daCunha2010,Iruskieta2013} and \cite{Hovy2010} indicates, a debate was carried out between annotators to solve disagreements. Thus, agreement was obtained for every case. This final corpus segmented was used as gold standard and will be available online for the scientific community. 
Table \ref{tab:gold} shows the gold standard statistics. 
As it can be observed in this table, the corpus includes 183 sentences; by contrast, it contains a higher number of discourse segments (280), which means that intra-sentence discourse segmentation is productive. 

\begin{table}[h]
\begin{center}
 \begin{tabular}{|l|r|c|c|r|}
 \hline
            & { \bf Total} & { \bf Longuest} & { \bf Shortest}  & {  \bf Average} \\
            &                           & { \bf text    } & {  \bf text   }  &    \\
 \hline
 {\bf  Num. of words}    &{ 4 676} & { 317}  & { 91}  & { 233.80}   \\
 {\bf  Num. of sentences}&{ 183}   & { 17}   & { 4}   & { 9.15}   \\
 {\bf  Num. of segments} &{ 280}   & { 24}   & { 8}   & { 14.00}   \\
 \hline
 \end{tabular}
\caption{Gold Standard statistics.}
\label{tab:gold}
\end{center}
\end{table}

\subsection{Evaluation}

We consider two baseline segmenters to compare our results:

\begin{itemize}
	\item {Baseline}$_1$: it inserts boundaries before coordinating conjunctions. 
	\item {Baseline}$_2$: it considers all complete sentences like discourse segments. This baseline will have a precision of 100\%, because all detected segments will be correct, since sentences are considered discourse segments.

\end{itemize}

The results obtained are shown in Table \ref{tab:results}. 
\begin{table}[h]
\begin{center}
 \begin{tabular}{|c|c|c|c|}
 \hline
 {\bf System} & {\bf F-Score} & {\bf Precision}  & {\bf Recall} \\
 \hline
 {\bf DiSegCAT}    &{ \bf 75\%}   & {  68\%}  & { \bf 85\%}    \\
 {\bf Baseline$_1$ } &{ 52\%}   & { 44\%}  & { 65\%}    \\
 {\bf Baseline$_2$ } &{ 18\%}   & {\bf 100\%} & { 10\%}    \\
 \hline
 \end{tabular}
\caption{Results of our experiment.}
\label{tab:results}
\end{center}
\end{table}
Differences between DiSegCAT and Baseline$_2$ scores are all significant based on 12-fold t-test with $p$-value $<0.05$.

Since there is not another discourse segmenter for Catalan we cannot compare our results with another system. This is a current situation when working on local languages.  We did try to combine DiSeg for Spanish with mainstream available translators form Spanish to Catalan.
But translation is an even more complex problem than discourse analysis and in the case of our corpus in Spanish, translators were not efficient with high rate of errors including over short multi-word expressions. 

However, we find that our results are similar to those obtained for other languages by using similar strategies of segmentation, such as for English (F-Score = 83\%) and Spanish (F-Score = 80\%).

As it can be observed in Table \ref{tab:results}, our system (DiSegCAT) obtains the best F-Score (75\%), in comparison with Baseline$_1$ (52\%) and Baseline$_2$ (18\%). As expected, Baseline$_2$ obtains 100\% of precision, since it considers sentences as segments; however, it obtains only 10\% of recall, since it does not detect intra-sentence segments. Baseline$_1$ has 44\% of precision and 65\% of recall. Although this baseline detects correctly some segments (because coordinated clauses can be also discourse segments), both results (precision and recall) are worst than the results obtained by DiSegCAT. These results mean that our algorithm outperforms baselines including linguistic information.

Regarding the results obtained by DiSegCAT, its performance is better for recall than for precision (85\% and 68\%, respectively). After obtaining these quantitative results, we have carried out a qualitative analysis in order to find different types of errors of the system. 

With respect to precision, the main error is related to coordination.  See for example the following segments, obtained automatically by DiSegCAT\footnote{English translation of examples has been extracted from the papers published by the authors in the journal \textsl{Apunts: Medicina de l'esport.}}:

\begin{verbatim}
[El nostre objectiu fou establir quins paràmetres antropomètrics]
[i de maduració es correlacionen amb el rendiment 
en rem-ergòmetre en una mostra de 114 adolescents d'ambdós sexes, 
sense experiència prèvia en rem.]
\end{verbatim}

{\small
\begin{verbatim}
[We aimed to establish which anthropometric]
[and maturity offset parameters correlate with 
rowing ergometer performance in a sample of 114 adolescent, 
rowing-inexperienced boys and girls.]
\end{verbatim}
}

Here, following our segmentation criteria and rules, the correct segmentation would be:

\begin{verbatim}
[El nostre objectiu fou establir quins paràmetres antropomètrics 
i de maduració es correlacionen amb el rendiment en rem-ergòmetre 
en una mostra de 114 adolescents d'ambdós sexes, sense experiència 
prèvia en rem.]
\end{verbatim}

The passage \textsl{quins [...] rem} is the direct object of the main verb (\textsl{fou establir}, ``was to establish'') of the sentence. Therefore, it should not be segmented. Nevertheless, this direct object includes a coordination that contains the conjunction \textsl{i} (``and'') and the finite verb \textsl{es} (``is''). 
Thus, the system segments after the conjunction, since one of the rules indicates that a passage written after this conjunction should be a discourse segment if it includes a finite verb. As it can be observed, in this case, the performance of this rule is not adequate. We should find a solution for this problem in the future. Coordination is also one of the main difficulties found in the performance of the discourse segmenter for Spanish \cite{daCunha2012}.

With regard to recall, the main problem is related to segments that are not explicitly marked in the text. See for example the following segment, obtained automatically by the system:

\begin{verbatim}
Té un cost baix, és massiva i de fàcil aplicació.
\end{verbatim}

{\small
\begin{verbatim}
It is low cost, it is massive and easy to use.
\end{verbatim}
}

The adequate segmentation of this passage should be:

\begin{verbatim}
[Té un cost baix]
[és massiva i de fàcil aplicació.]
\end{verbatim}

The second segment includes a finite verb, but there is no specific mark indicating that it is a discourse segment (the comma is not included in our rules as a boundary mark, since it would over-generate discourse segments). In the future, we plan to study strategies to solve this limitation, although it is a difficult issue.

\section{Conclusions and Future Work}
\label{sec:conclusion}

This paper presents the first discourse segmentation system for Catalan based on RST.
The segmenter uses simple linguistic rules and promising results have been obtained in our experiments. 
This system could be used in different tasks in the context of NLP. 
For example, segmentation of (too) long sentences which are
difficult to parse; elimination of text segments in Sentence Compression and Automatic Text Summarization; Alignment of text in different languages for Machine Translation, etc.
Also, the segmenter can be the basis for further development of an automatic discourse parsing system for Catalan, since this tool does not exist so far.

The methodology used to develop this segmenter for Catalan is very similar to that used for Spanish.
In fact, the good results obtained show us that this methodology is probably valid in general for Romance languages.
The future experiments will be conducted on this line. 
We also plan to expand the size of the corpus, adding texts of other genres (such as news reports) and other domains (such as Linguistics or Economy).
 
\section*{Acknowledgements}
This work has been partially supported by a Ram\'on y Cajal research contract (RYC-2014-16935) and the research project APLE 2 (FFI2009-12188-C05-01) of the Institute for Applied Linguistics (IULA).

%
%
\bibliographystyle{splncs}
\bibliography{biblio}
\end{document}